\documentclass[10pt,twocolumn,letterpaper]{article}

\usepackage{cvpr}              % To produce the CAMERA-READY version
\newcommand{\smallsection}[1]{\noindent\underline{\smash{\textbf{#1:}}}}

\usepackage[accsupp]{axessibility}
\usepackage{amsmath,amssymb,amsthm}
\usepackage{xcolor}
\usepackage{comment}
\usepackage{xspace}
\usepackage{enumitem}
\usepackage[ruled,linesnumbered,noend]{algorithm2e}
\usepackage{multirow}
\usepackage{booktabs}
\usepackage{url}
\usepackage{makecell}
\usepackage{adjustbox}
\usepackage{dsfont}
\usepackage{bbm}
\usepackage{colortbl}

\usepackage{mathtools}
\usepackage{bigstrut}
\usepackage{xurl}
\usepackage{pifont}
\definecolor{newgreen}{rgb}{0.0, 0.5, 0.0}
\definecolor{newred}{rgb}{0.81,0.1,0.26}
\newcommand{\yes}{\textcolor{newgreen}{\textbf{\ding{52}}}}
\newcommand{\no}{\textcolor{newred}{\textbf{\ding{55}}}}

\newcommand{\GCN}{GatedGCN\xspace}
\newcommand{\DAGNN}{DAGNN\xspace}
\newcommand{\DAGFo}{DAGFormer\xspace}
\newcommand{\GPS}{GraphGPS\xspace}
\newcommand{\TAGATES}{TA-GATES\xspace}
\newcommand{\NARFo}{NAR-Former\xspace}

\newcommand\blue[1]{\textcolor{blue}{#1}}

\definecolor{peace}{RGB}{228, 26, 28}
\definecolor{love}{RGB}{55, 126, 184}
\definecolor{joy}{RGB}{77, 175, 74}
\definecolor{kindness}{RGB}{152, 78, 163}
\definecolor{crimson}{rgb}{0.86, 0.08, 0.24}
\definecolor{sunwooyellow}{rgb}{1.0, 1.0, 0.0}
\definecolor{sunwoogreen}{rgb}{0.66, 0.89, 0.63}
\definecolor{newyellow}{rgb}{1.0,0.8,0.02}
\definecolor{newgreen}{rgb}{0,0.7,0.0}

\newcommand{\method}{\textsc{FlowerFormer}\xspace}
\newcommand{\flayer}{\textsc{Flower}\xspace}

\newcommand{\std}{\scriptsize}

\newcommand{\best}{\cellcolor{sunwoogreen}}  %{0.9}
\newcommand{\secb}{\cellcolor{white}}  %{0.9}

\newcommand\green[1]{\textcolor{newgreen}{#1}}

\def\thickhline{\noalign{\hrule height1.2pt}}

\def\paperID{17186} % *** Enter the Paper ID here
\def\confName{CVPR}
\def\confYear{2024}

\definecolor{cvprblue}{rgb}{0.21,0.49,0.74}
\usepackage[pagebackref,breaklinks,colorlinks,citecolor=cvprblue]{hyperref}
\title{FlowerFormer: Empowering Neural Architecture Encoding \\ using a Flow-aware Graph Transformer}

\author{Dongyeong Hwang \quad Hyunju Kim \quad Sunwoo Kim \quad Kijung Shin\\
Kim Jaechul Graduate School of AI, KAIST, Seoul, Republic of Korea\\
{\tt\small \{dy.hwang, hyunju.kim, kswoo97, kijungs\}@kaist.ac.kr}
}
\begin{document}
\maketitle

\begin{abstract}
%In the rapidly evolving field of deep learning, a wide array of models has emerged, each with architectures that exhibit varying degrees of effectiveness depending on the specific dataset and task they are applied to. 
The success of a specific neural network architecture is closely tied to the dataset and task it tackles; there is no one-size-fits-all solution.
Thus, considerable efforts have been made to quickly and accurately estimate the performances of neural architectures, without full training or evaluation, for given tasks and datasets.
Neural architecture encoding has played a crucial role in the estimation, and graph-based methods, which treat an architecture as a graph, have shown prominent performance.
For enhanced representation learning of neural architectures, we introduce \method, a powerful graph transformer that incorporates the information flows within a neural architecture.
\method consists of two key components: (a) bidirectional asynchronous message passing, inspired by the flows; (b) global attention built on flow-based masking.
%Our tests show \method surpasses other models, highlighting the importance of understanding information flow in neural architectures. 
Our extensive experiments demonstrate the superiority of \method over existing neural encoding methods, and its effectiveness extends beyond computer vision models to include graph neural networks and auto speech recognition models.
%highlighting the importance of modeling the information flow of neural architectures. 
%Moreover, the efficacy of \method 
%as evidenced by the outstanding performance of \method in benchmark datasets of other domains, including graph neural networks and auto speech recognition models,
Our code is available at \url{http://github.com/y0ngjaenius/CVPR2024_FLOWERFormer}.
%goes beyond vision models. 
%It also performs well in the performance prediction of graph neural networks and automatic speech recognition architectures, showing its versatility and setting a new benchmark for predicting performance. Code is available at \url{http://anonymous.4open.science/r/FLOWERFormer/}.

%The success of a specific neural network architecture is closely tied to the dataset and task it tackles; there is no one-size-fits-all solution. Thus, considerable efforts have been made to quickly and accurately estimate the performances of neural architectures, without full training or evaluation, for given tasks and datasets. Neural architecture encoding has played a crucial role in the estimation, and graph-based methods, which treat an architecture as a graph, have shown prominent performance. For enhanced representation learning of neural architectures, we introduce FlowerFormer, a powerful Graph Transformer that incorporates the information flows within a neural architecture. FlowerFormer consists of two key components: (a) bidirectional asynchronous message passing, inspired by the flows, and (b) global attention built on flow-based masking. Our extensive experiments demonstrate the superiority of FlowerFormer over existing neural encoding methods, and the effectiveness extends beyond computer vision models to include graph neural networks and auto speech recognition models.

\end{abstract}

\section{Introduction}
\label{sec:intro}

While deep learning models have demonstrated their efficacy across various applications, 
the performance of a specific neural architecture heavily depends on specific downstream tasks and datasets employed.
As a result, numerous neural architectures have been developed~\cite{he2016deep, huang2017densely}.
%Then, given a dataset and a task, how can we rapidly and accurately estimate the performance of a neural architecture?

In response to this dependency, significant efforts have been made to rapidly and accurately predict the performances of neural architectures for given tasks and datasets.
This endeavor is crucial because exhaustively training and/or evaluating many candidate neural architectures is an expensive process.  
To this end, researchers have primarily employed machine learning techniques~\citep{luo2018neural, chen2021not}.

Especially, various neural architecture encoding methods have been proposed since obtaining an accurate representation of each architecture plays a crucial role in the estimation process.
Their focus has mainly revolved around (a) transforming input neural architectures to appropriate data structures~\citep{luo2020semi, white2021bananas} and (b) applying representation-learning models to the transformed structures~\citep{chen2021contrastive,white2021powerful}.

Some have treated neural architectures as graphs and applied graph representation learning. They, however, share some limitations. For instance, their basic message-passing mechanisms oversimplify neural-architecture characteristics~\citep{shi2020bridging, wen2020neural} and may suffer from over-smoothing~\citep{oono2019graph}, over-squashing~\citep{alon2020bottleneck}, or limited expressiveness~\citep{rampavsek2022recipe}.

%Mitigating such issues, Graph Transformers (GTs) have shown remarkable performance as a graph encoder, especially in graph classification~\citep{ying2021transformers, hussain2022global} and regression~\citep{chen2023graph,ma2023graph} tasks.
Graph Transformers (GTs), when incorporated with adequate information, are recognized for enhancing basic message passing, making them effective in various graph classification~\citep{ying2021transformers, hussain2022global} and regression~\citep{chen2023graph,ma2023graph} tasks.
One strength of GTs lies in their global attention mechanisms~\citep{vaswani2017attention}, where all nodes in an input graph contribute directly to forming the representation for each individual node. 

% However, it is crucial to integrate relevant topological or external information to ensure the relevance of attention scores and, consequently, the effectiveness of GTs.
% For example, \citet{niu2022chemistry} incorporate the characteristics of molecule graphs into GTs using motif-based spatial embedding. 
However, without integrating relevant topological or external information of input graphs, the relevance of attention scores, and thus the effectiveness of GTs, might be impaired.
For example, \citet{niu2022chemistry} showed the essentiality of using motif-based spatial embedding to incorporate the characteristics of molecule graphs into GTs.

In this work, we propose \method (\textbf{\underline{Flow}}-awar\textbf{\underline{e}} g\textbf{\underline{r}}aph trans\textbf{\underline{former}}), a GT model specialized in capturing \textit{information flows} within neural architectures, as illustrated in \cref{fig:nnflow}.
% Capturing the flows is essential because a neural architecture derives its fundamental properties through the flows of forward and backward propagation.
The information flows of a neural architecture contain the characteristics of both forward and backward propagations of the architecture, and thus describe its fundamental properties.
\method includes two core modules: the \textit{flow encode} module and the \textit{flow-aware global attention} module.
The former conducts bidirectional asynchronous message passing, imitating the forward and backward propagations within the input neural architecture.
The latter applies global attention with masking schemes based on the flow-based dependencies between nodes.

Our extensive experiments on neural architecture performance prediction, conducted using five benchmark datasets,
validate the superiority of \method over state-of-the-art neural encoding models~\citep{ning2022ta,yi2023nar}.
The results highlight the effectiveness of incorporating flows into GTs. Our contributions are summarized as follows:

\begin{figure}[t]
    %\vspace{-2mm}
    \centering
    \includegraphics[width=0.9\linewidth]{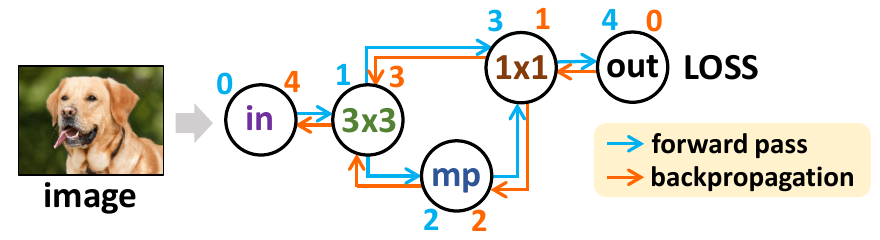} \\
    % \vspace{-1mm}
    \caption{Information flows within an example neural architecture from the NAS-Bench-101 benchmark~\citep{ying2019bench}. 
    The architecture is represented as a directed graph where each node corresponds to an operation, and
     the topological structure of the graph encodes the sequence in which these operations are performed.
    For instance, the `$1\times1$' (convolution) operation is executed only after the `$3\times3$' (convolution) and `mp' (max pooling) operations have been completed.
    The forward pass, depicted by blue arrows, is followed by the backpropagation of the loss, depicted by orange arrows.
    The number displayed above each node indicates the processing order within each flow.}
    % \vspace{-2mm}
    \label{fig:nnflow}
\end{figure}

\begin{itemize}
    \item We propose \method, a flow-aware GT-based neural architecture encoding model.
    To our best knowledge, \method is the first GT model specifically designed to capture flows.

    \item \method outperforms six baseline architectures, including the most recent ones~\citep{ning2022ta,yi2023nar}, by a substantial margin across three benchmark datasets in the computer vision domain.
    Specifically, in predicting the performance of neural architectures, it outperforms the top-performing baseline method by a margin of up to 4.38\% in Kendall's Tau. Additionally, through ablation studies, we justify the design choices made in \method. 

    \item Beyond computer vision neural architectures, \method also excels at performance prediction for graph neural networks and auto speech recognition architectures. In the benchmarks for these architectures, \method achieves performance gains of up to 4.41\% in Kendall's Tau over baseline models.
\end{itemize}

 Our code is available at \url{http://github.com/y0ngjaenius/CVPR2024_FLOWERFormer}.

\section{Related work}
\label{sec:related}
In this section, we briefly review related studies in neural architecture encoding and graph transformers (GTs).

%-------------------------------------------------------------------------
    \subsection{Neural architecture encoding}
   \label{related:neural}
   
    Neural architecture encoding~\cite{luo2018neural, liu2018progressive, wang2019alphax, luo2020semi, white2021bananas}, which aims to learn representations of neural architectures, has gained considerable attention due to its significant downstream tasks, such as performance prediction (i.e., the prediction of task- and data-specific performance for given architectures without full training or evaluation). 
    %The most well-known approach for predicting the performance of an architecture is to obtain its representation with high quality. 
    %Consequently, a variety of approaches, including MLP, and LSTM, have been established.
    
    One popular class of approaches is graph-based, modeling neural architectures as graphs and using graph neural networks~\cite{kipf2016semi} for representation learning. These approaches have also introduced topology-based graph similarity and operation-specific embeddings~\cite{cheng2021nasgem, chatzianastasis2021graph}.
    
    %propose graph encoding schemes that take into account graph similarity based on their topology or acquire operation embeddings within the architectural framework.
    
    Another significant approach aims to obtain representations that mimic the forward and/or backward passes within neural architectures.
    For instance, GATES~\cite{ning2020generic} updates operation embeddings by mimicking the application of operations to information (which is also represented as a vector) and thus effectively replicating the forward-pass of convolution operations. 
    Another method, \TAGATES~\cite{ning2022ta}, simulates an iterative process involving both forward and backward passes, with specialized handling for specific operations, e.g., skip-connections.
    However, these methods focus on flows only at a local level, by simulating a series of local operations, and may overlook a global-level perspective. %Its advanced version, 

    Transformer-based models~\cite{yan2021cate, lu2021tnasp} are capable of capturing global-level perspectives through attention mechanisms. \NARFo~\cite{yi2023nar}, a multi-stage fusion transformer, is one of the state-of-the-art methods for predicting neural architecture performance.
    They (1) represent a neural architecture as a sequence of operations to employ a transformer-based model and (2) leverage multiple valid sequences from the same architecture for augmentation. 
 
%    \blue{On the other hand, transformer-based models have been developed to predict the performance, where \NARFo~\cite{yi2023nar} which introduces a multi-stage fusion transformer is recognized as the SOTA. They conceptualize the neural network as a sequence of operations and employ a transformer-based model for its encoding. Additionally, they introduce an information flow consistency augmentation. The augmentation utilizes graph isomorphism, using the fact that graphs are invariant under different node numberings. They avoid conflicts between node numbering and the computational processing order of the neural network.}
    
    In this work, we unify all three dimensions—graph learning, flow modeling, and global attention—by introducing a novel flow-aware GT, marking the first instance of such integration to the best of our knowledge.

\begin{figure*}[t!]
    \vspace{-2mm}
  \centering
    \includegraphics[width=0.9\linewidth]{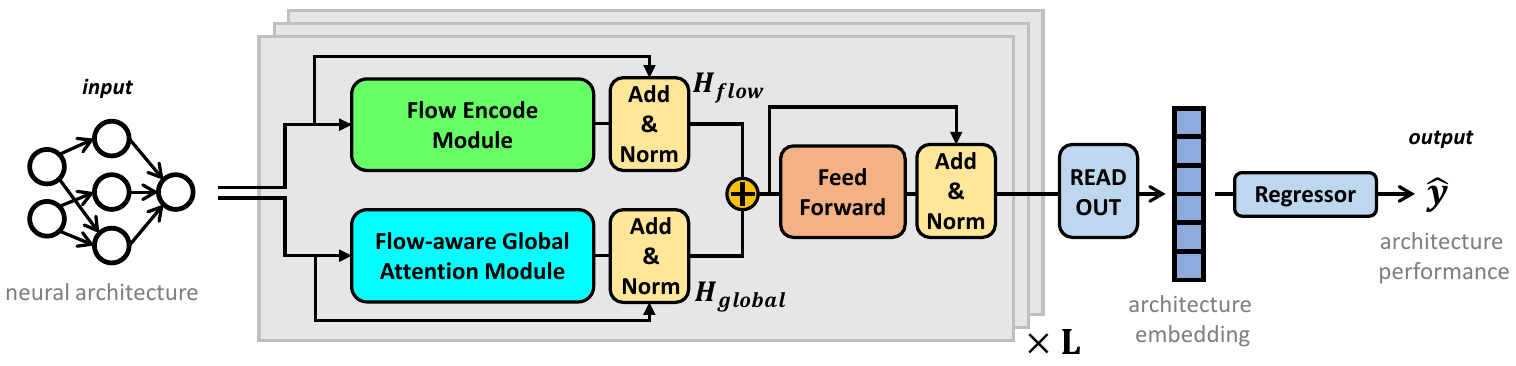} \\
    % \vspace{-1mm}
    \caption{Overview of proposed \method, which contains two key modules in each of its layers: the \textit{flow encode} module and the \textit{flow-aware global attention} module. The \textit{flow encode} module performs bidirectional asynchronous message passing, inspired by forward and backward passes, to produce a node embedding matrix $H_{\text{flow}}$. The \textit{flow-aware global attention} module computes attention with a flow-based masking scheme to yield another node embedding matrix $H_{\text{global}}$. 
    These two embedding matrices, $H_{\text{flow}}$ and $H_{\text{global}}$, are combined and then projected to produce updated node embeddings at each layer. This process is iterated over $L$ layers, and the output node embeddings are aggregated to form the final architecture embedding, which is fed into a regressor for performance prediction.
    }
    % \vspace{-2mm}
    \label{fig:overview}
\end{figure*}

    \subsection{Graph transformers (GTs)}
    \label{related:gt}

    Graph transformers (GTs) \cite{dwivedi2020generalization, kreuzer2021rethinking, ying2021transformers, shi2022benchmarking, hussain2022global, rampavsek2022recipe} apply global (i.e., graph-level) attention between all node pairs. Recently, GTs show remarkable performance in various graph-level tasks, including molecular property prediction~\cite{rong2020self, jiang2023pharmacophoric}, image classification~\cite{nguyen2021modular, zheng2022graph}, and human interaction recognition~\cite{pang2022igformer}.  

%    The core insight of graph transforemrs (GTs) is to utilize global attention~\citep{vaswani2017attention} between all pairs of nodes. Due to this, GTs~\cite{dwivedi2020generalization, kreuzer2021rethinking, ying2021transformers, shi2022benchmarking, hussain2022global, rampavsek2022recipe} have demonstrated success in addressing graph-level tasks including molecular property prediction~\cite{rong2020self, jiang2023pharmacophoric}, image classification~\cite{nguyen2021modular, zheng2022graph}, and human interaction recognition~\cite{pang2022igformer}. 

    To further improve their effectiveness, global attention is often supplemented with topological and/or external information. The information includes eigenvectors of adjacency and Laplacian matrices~\cite{shi2022benchmarking, kreuzer2021rethinking} and pair-wise node similarity derived from shortest paths, diffusion kernels, random walks, etc~\cite{kreuzer2021rethinking, ying2021transformers, mialon2021graphit}.

   Some GTs are tailored for specific types of graphs. For molecular graphs, where motifs play key roles,
   \citet{niu2022chemistry} employ motif-based spatial embeddings in a GT.
   \DAGFo~\cite{luo2023transformers} is designed for directed acyclic graphs (DAGs) and incorporates depth-based positional encodings and reachability-based attention. Note that \DAGFo is designed for general DAGs, and it is not optimized for encoding neural architectures, especially in capturing architecture-specific flows.

   %Note that \DAGFo is a model designed for general DAGs and is not specifically optimized for learning the flow which is crucial to represent neural architectures. Also, 

% \section{Continuous Relaxation of Rumor Blocking}
% \label{sec:concepts}
% \input{030notation}

\section{Proposed method: \method}
\label{sec:method}
In this section, we present \textbf{\method} (\textbf{\underline{Flow}}-awar\textbf{\underline{e}} g\textbf{\underline{r}}aph trans\textbf{\underline{former}}), a graph transformer model designed to capture information flows within an input neural architecture. 
First, we provide the motivation behind \method in \cref{method:motivation}.
Then, we describe how an input neural architecture is represented as a graph in~\cref{method:input}.
After that, we elaborate on how \method learns the representation of the neural architecture graph.
Specifically, we describe two core modules of \method, collectively referred to as  \flayer, in \cref{method:layer}.
Lastly, we present the overall framework (refer to \cref{fig:overview}) in \cref{method:overall}.

\subsection{Motivation of capturing information flows}\label{method:motivation}
Despite the remarkable success of Graph Transformers (GTs) in various graph-level tasks, including graph classification~\citep{ying2021transformers, hussain2022global} and regression~\citep{chen2023graph,ma2023graph}, their application for encoding neural architectures has received relatively limited attention.
% The effectiveness of GTs originates not only from their global attention mechanism between all pairs of nodes but also from additional design choices for accurately capturing the underlying characteristics of input graphs. 
Existing applications of GTs suggest that additional design choices for accurately capturing the underlying characteristics of input graphs (on top of global attention mechanism between all pairs of nodes) are essential for the effectiveness of GTs.
Refer to~\cref{related:gt} for some examples.

%through their global feature capture, their effectiveness in neural architecture performance prediction remains limited. 
%Given that Graph Transformers predominantly utilize node features, their effectiveness necessitates the provision of supplementary information. Determining the appropriate characteristics of the neural network to incorporate remains a complex task.}

In this work, we focus on a crucial aspect: capturing \textit{information flows} within neural architectures (i.e., input graphs). 
Information flows include both the forward pass of data and the backpropagation of gradients.
Hence, it is essential to capture information flows for incorporating how neural architectures are trained and conduct inference into their embeddings (i.e., the encoded neural architectures).

%ow it is being trained and makes an inference. This mechanism can be expressed by \textit{information flows} within a neural architecture. Specifically, the flows depict the forward pass of data and the backpropagation of gradients. By better capturing this information flow, GT can learn better embedding of the neural network.

%\subsection{Constructing Input Graph}~\label{method:input}
%In this subsection, we describe the input modeling for \method, transforming an input neural architecture to a graph.

\subsection{Input modeling}\label{method:input}

% \crim{\textit{Information flow} is a representative underlying characteristic of a neural architecture graph. 
% Training of neural networks mainly adopts a forward-backward propagation scheme, and this propagated. [Justification]}

%highlighting the sequential nature of neural networks. 
%Data traverses through operations following a topologically ordered sequence. Understanding flow at the local connection level is vital, providing critical insight for Graph Transformers. It justifies the inclusion of flow as essential additional information to enhance the performance prediction ability of GT.}

\begin{figure}[t]
    %\vspace{-2mm}
    \centering
    \includegraphics[width=0.95\linewidth]{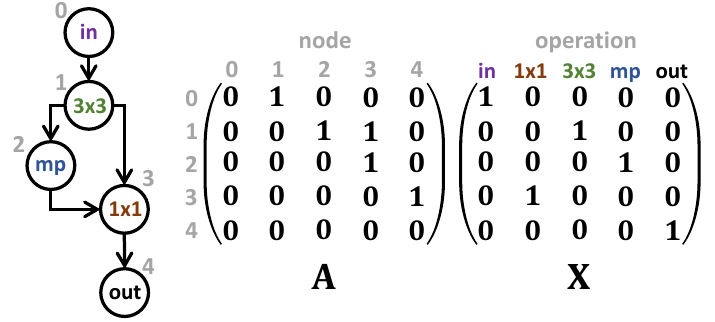} \\
    % \vspace{-1mm}
    \caption{An example neural architecture from the NAS-Bench-101 dataset, represented as a directed acyclic graph (DAG), and its adjacency matrix $A$. Each column of the node feature matrix $X$ corresponds to a specific operation, and each row in $X$ is a one-hot vector indicating the type of operation associated with the corresponding node.}
    % \vspace{-2mm}
    \label{fig:graph}
\end{figure}

We represent a given neural architecture as a directed acyclic graph (DAG), with each node representing an operation (e.g., pooling or convolution).
Each directional edge between two nodes indicates the information flow between the corresponding operations, aligning with the direction of data propagation during the forward pass. An illustrative example can be found on the left-hand side of Figure \ref{fig:graph}.

% Considering the propagation scheme of a neural network, expressing a neural architecture as a Directed Acyclic Graph (DAG) is intuitive and appropriate. 
% Nodes of the DAG correspond to operations, such as pooling and convolution. 
% Directional edges correspond to the data flow between these operations, where their directions indicate the input flow within the neural network.
% An example is depicted in the left-hand side figure of \cref{fig:graph}.

We denote the graph representation of a neural architecture by $G=(A, X)$, a tuple of an adjacency matrix $A\in \{0,1\}^{N\times N}$ and a node (i.e., operation) feature matrix $X \in \{0,1\}^{N\times D}$, where $N$ is the number of nodes and $D$ is the number of operations. 
The adjacency matrix encodes direct connections between node pairs in a graph. Its binary entries indicate whether a directional edge exists between each pair of nodes. Specifically, the $(i,j)$-th entry of $A$ is set to 1 if there is a directed edge from the $i$-th node (denoted as $v_i$) to the $j$-th node (denoted as $v_j$), and 0 otherwise. 
Each node is associated with a one-hot feature vector representing its corresponding operation, and these vectors are stacked vertically to form the node feature matrix $X$. Refer to~\cref{fig:graph} for an example.

% The input modeling scheme employed by \method stands in contrast to state-of-the-art neural encoding methods, which rely on complex modeling and preprocessing steps, such as the specialized treatment of specific operations~\cite{ning2022ta} and isomorphic augmentations~\cite{yi2023nar} (refer to \cref{related:neural}).
% The empirical superiority of \method (refer to \cref{sec:experiments}) despite its simple input modeling is attributed to our novel graph-transformer architecture, which is described in the following subsection.
With our general input modeling scheme, \method is readily applicable to different domains and neural architectures without such additional modelings or steps.
By contrast, state-of-the-art neural encoding methods often rely on complex modelings and/or preprocessing steps, such as the specialized treatment of specific operations~\cite{ning2022ta} and isomorphic augmentations~\cite{yi2023nar} (refer to \cref{related:neural}).
The empirical superiority of \method (refer to \cref{sec:experiments}) despite its straightforward (yet elegant) input modeling is attributed to our novel flow-aware GT architecture, which is described in the following subsection.

\subsection{\textbf{\flayer} layers}~\label{method:layer}
In this section, we introduce \flayer layers, the basic units of \method. %It aims to incorporate the information flows within a neural architecture. 
%To learn the information flow within the architecture, 
A \flayer layer consists of two core components: the \textit{flow encode} module and the \textit{flow-aware global attention} module.
The flow encode module is a message-passing neural network (MPNN) that asynchronously passes messages in the forward and then the backward orders.
The flow-aware global attention module is a self-attention module based on a flow-aware masking scheme. 
The outputs of the flow encode module and the flow-aware global attention module are node embedding matrices, denoted as  $H^{(l)}_{flow}\in\mathbb{R}^{N\times d}$ and $H^{(l)}_{global}\in\mathbb{R}^{N\times d}$, respectively, for the $l$-th \flayer layer.
Below, we provide a detailed explanation of each module.
%as output node embedding matrices after the flow encode module and flow-aware global attention module in the $l$-th layer.
%We use $H\in\mathbb{R}^{N\times d}$ for an embedding matrix. 
\begin{algorithm}[t]
    \small
    \caption{\text{Flow encode module}}\label{algo:flowenc}
    \SetKwInput{KwInput}{Input}
    \SetKwInput{KwOutput}{Output}
    \KwInput{(1) $G =  (A, X)$: an input neural architecture \\
    \quad\quad\quad (2) $H$: an input node embedding matrix \\
    }
    \KwOutput{$H$: updated node embedding matrix }    

    \vspace{1mm}
    \blue{${/}^{*}$ step 1. topological sorting ${}^{*}/$} \\
    $\mathcal{T}^G\leftarrow \text{topological generations of } G$\label{algo:flowenc:init} \\
    \vspace{1mm}

    \blue{${/}^{*}$ step 2. asynchronous forward message passing ${}^{*}/$} \\
    \For{$k = 1,\dots, |\mathcal{T}^G|$ \label{algo:flowenc:forward:start}}{
        \For{$v_j\in T^G_k$}{
            $h_j\leftarrow \operatorname{Comb}(h_j, \operatorname{Agg}\{m_e(h_j, h_i): A_{ij}=1\})$\label{algo:flowenc:forward} \label{algo:flowenc:forward:end}
        }
    }
    %$\Tilde{G} \leftarrow \text{reversed DAG}$\\
    %$\mathcal{T}^{\Tilde{G}}\leftarrow \text{topological generations of } \Tilde{G}$ \\

    \vspace{1mm}
    \blue{${/}^{*}$ step 3. asynchronous backward message passing ${}^{*}/$} \\
    \For{$k = |\mathcal{T}^{G}|, \dots, 1$ \label{algo:flowenc:backward:start}}{
        \For{$v_j\in T^{G}_k$}{
            $h_j\leftarrow \text{Comb}(h_j, \text{Agg}\{m_e(h_j, h_i): A_{ji}=1\})$\label{algo:flowenc:backward} \label{algo:flowenc:backward:end}
        }
    }
 \Return $H$
 
\end{algorithm}

\subsubsection{Flow encode module}

As discussed in \cref{method:motivation}, we aim to enable a GT to capture the crucial aspect of neural architectures—\textit{information flows}. To this end, the flow encode module conducts both asynchronous forward and backward message passing,  resembling the forward pass (i.e., inference) and backpropagation (i.e., training) of neural architectures, respectively.
These message-passing procedures are carried out in the (reversed) topological order in the input neural architecture graph, leading to updated node embeddings.

%Within neural architecture, data progresses sequentially following the topological order of operations. Additionally, neural architecture features a learning flow that engages backpropagation following the forward pass of data~\citep{ning2022ta}. Our design reflects these characteristics through tailored elements for each functional aspect.

Pseudocode of the flow encode module is presented in~\cref{algo:flowenc}.
It includes topological sorting, forward message passing, and backward message passing, in order, and each of these components is described below.

\begin{figure}[t]
    %\vspace{-2mm}
    \centering
    \includegraphics[width=0.7\linewidth]{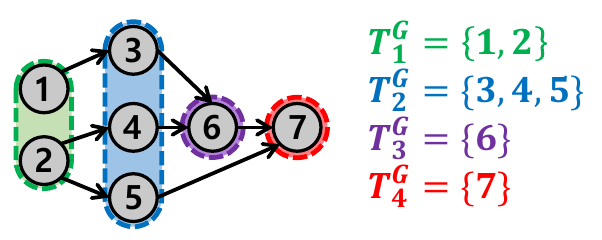} \\
    % \vspace{-1mm}
    \caption{Example topological generations. 
    %\blue{We delete a set of nodes and their incident edges sequentially of those without incoming edges.}
    Nodes 1 and 2 are devoid of incoming edges, and thus they constitute the first topological generation $T^{G}_{1}$. 
    Upon removal of nodes 1 and 2, nodes 3, 4, and 5 no longer have incoming edges, and thus they compose the second generation $T^{G}_{2}$.
    Subsequently, nodes 6 and 7 form the third and fourth generations, respectively.}
    % \vspace{-2mm}
    \label{fig:top}
\end{figure}

\smallsection{Topological sorting (\cref{algo:flowenc:init})}
The first step is to divide nodes (i.e., operations) into topological generations.
Recall that neural-architecture graphs are directed acyclic graphs (DAGs).
Given a DAG $G$, its first topological generation, denoted as $T^G_1$, comprises the nodes without incoming edges in $G$.
Then, for each $k>1$, the $k$-th topological generation $T^G_k$ comprises the nodes without incoming edges when all preceding generations are removed from $G$.
The set of non-empty topological generations is denoted as $\mathcal{T}^G:=\{T_1^G, \dots, T_{|\mathcal{T}^G|}^G\}$. Refer to \cref{fig:top} for an example.

These topological generations are closely related to the data flow within a neural architecture. For the operations (i.e., nodes) in each generation to be executed, all operations in the preceding generations need to be complete.
Conversely, during the process of backpropagation, gradients flow from subsequent generations to preceding generations.

%To express the neural architectures's processing order, we utilize a variant of topological sorting. Each step consists of removing all nodes that have no in-neighbors~\footnote{The in-neighbors of $v_j$, is a set of nodes $v_i$ such that $A_{ij}=1$.} and edges incident to them, instead of deleting a node one by one. 
%Then the topological generations of $G$, denoted as $\mathcal{T}^G$, is a class of $T^G_1, \dots, T^G_{|\mathcal{T}^G|}$, i.e., $\mathcal{T}^G:=\{T_1^G, \dots, T_{|\mathcal{T}^G|}^G\}$. Here, $T^G_k$ represents a set of nodes that are deleted during the $k$-th step.

\smallsection{Forward message passing (\cref{algo:flowenc:forward:start}-\cref{algo:flowenc:forward:end})}
During the forward message passing step, node embeddings are updated asynchronously, following the order of the topological generations, akin to the forward pass within neural architectures.
For each node $v_j$, its embedding $h_j$ (i.e., the $j$-th row vector of $H$) is updated by the following three steps: (1) computing the message $m_e(h_j, h_i)$ for each incoming neighbor $v_i$, (2) aggregating these messages, and (3) combining the result with the current embedding $h_j$ (\cref{algo:flowenc:forward}).
Note that the embeddings of all incoming neighbors, which belong to preceding generations, have already been updated by the time message calculation occurs.
Also note that this differs from conventional synchronous graph message passing, where all node embeddings are updated simultaneously based on their input embeddings.

%To depict the sequential data processing in a neural architecture, we adopt asynchronous message passing. Note that, message passing in graph neural networks is conventionally synchronous, meaning that updates across the graph are performed simultaneously. Each node embedding is updated in order of the topological generations.
%For a node $v_j$, we obtain its node embedding $h_j$, the $j$-th row vector of $H$, by following three steps: (1) calculate messages $m_e(h_j, h_i)$ for every $v_i$ such that $A_{ij}=1$ (in-neighbors), (2) aggregate them, and (3) combine them with $h_j$ (\cref{algo:flowenc:forward}).

\begin{figure}[t]
    \vspace{-2mm}
    \centering
    \includegraphics[width=0.90\linewidth]{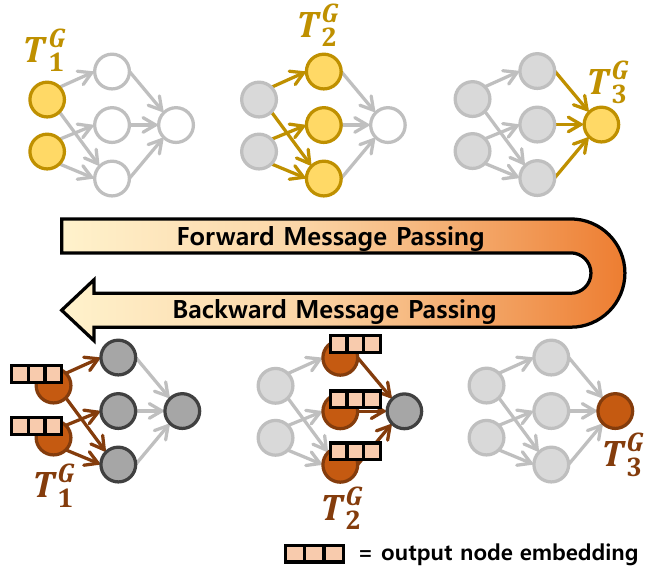} \\
    % \vspace{-1mm}
    \caption{Flow encode module. During forward message passing, node embeddings are updated following the order of topological generations. Conversely, during backward message passing, node embeddings are updated in the reverse order of the generations.}
    % \vspace{-2mm}
    \label{fig:message_passing}
\end{figure}

In our implementation, we use the sum aggregation as the $\operatorname{Agg}$ function. As $m_e$ and $\operatorname{Comb}$, we adopt the message function and the combine operator used in~\citep{thost2021directed}, as follows:
\begin{align}
    & m_e(h_j, h_i) = \operatorname{softmax}(w^{\top}_1 h_j + w^{\top}_2 h_i) h_i, \label{eq:message}\\
    & \operatorname{msg}_j = \Sigma_{i: A_{ij}=1} m_e(h_j, h_i), \label{eq:agg}\\
    & \operatorname{Comb}(h_j, \operatorname{msg}_j) = \operatorname{GRU}(h_j, \operatorname{msg}_j), \label{eq:comb}
\end{align}
\noindent
where $w_1\in \mathbb{R}^d$ and $w_2\in \mathbb{R}^d$ are learnable parameters.

\smallsection{Backward message passing (\cref{algo:flowenc:backward:start}-\cref{algo:flowenc:backward:end})}
After the forward message passing step, we further update node embeddings through backward message passing, which resembles the process of backpropagation. This aligns with the standard practice in neural architecture training, where backpropagation typically occurs after the forward pass for loss computation.

During the backward message passing step, node embeddings are updated asynchronously, following the reverse order of the topological generations.
For each node $v_j$, the messages from its outgoing neighbors (rather than incoming neighbors) are computed and then aggregated (\cref{algo:flowenc:backward:end}). %The aggregated result is combined with the current embedding for its update. 
The other details remain consistent with those of the forward message passing in to \cref{eq:message}, \cref{eq:agg}, and \cref{eq:comb}.

\smallsection{Outputs}
We denote 
the output node-embedding matrix of the flow encode module in the $\ell$-th \flayer layer, as $H^{(\ell)}_{flow}\in\mathbb{R}^{N\times d}$. That is,
\begin{align}
    H^{(\ell)}_{flow} &= \operatorname{FlowEncoder}(G, H^{(\ell-1)}). \label{eq:flow}
\end{align}
Here, $H^{(\ell-1)}\in\mathbb{R}^{N\times d}$ is the input node-embedding matrix obtained in layer-$(\ell - 1)$, the previous layer (\cref{eq:H_ell}).
%corresponds to the final output of the $(l-1)$-th \flayer layer for each $l>1$, and to $X$ when $l=1$.}

% In addition, our scheme (1) processes forward message passing from the $T^G_1$ to $T^G_{|\mathcal{T}^G|}$ and then (2) utilizes backward message passing to leverage flow information of backpropagation. We denote a reversed DAG of $G$ as $\tilde{G}$, where the direction of every edge becomes opposite so that $\tilde{G}=(A^T, X)$.
% Backward message passing flows on the reversed DAG $\tilde{G}$ with the opposite direction, which ensures that the node embedding update proceeds in alignment with the neural network's computation flow. This is summarized as follows:

\subsubsection{Flow-aware global attention module}
\begin{figure}[t]
    \vspace{-2mm}
    \centering
    \includegraphics[width=0.99\linewidth]{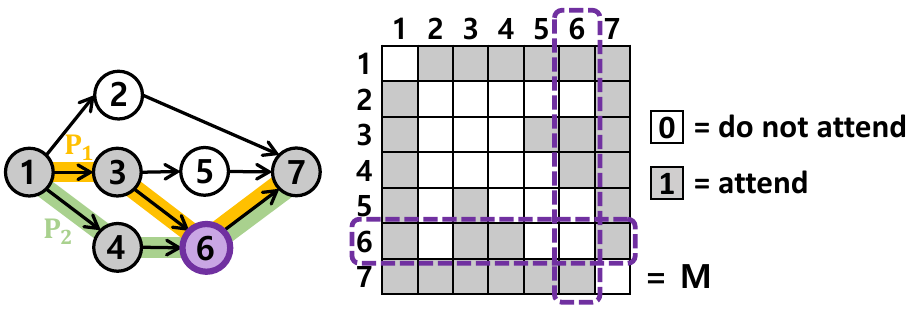} \\
    % \vspace{-1mm}
    \caption{An example mask matrix $M$. 
    Node $6$ attends exclusively to nodes that appear in any path involving the node ($P_1$ and $P_2$).
    Nodes 1, 3, and 7 appear in $P_1$, and nodes 1, 4, and 7 appear in $P_2$; thus node $6$ attends only to 1, 3, 4, and 7, as indicated by $M$.
    }
    
 %   \blue{we only compute attention for a node 6 with nodes on those paths (nodes 1,3,7 on $P_1$ and nodes 1,4,7 on $P_2$).}
 %   \red{[KJ: TODO]}}
    % \vspace{-2mm}
    \label{fig:attention}
\end{figure}

The flow-aware global attention module is designed to capture graph-level (i.e., architecture-level) characteristics, complementing the flow encode module which primarily focuses on local-level flows between directly connected operations.
To this end, we employ a global attention mechanism of GTs; moreover, to accurately reflect the flows within architectures, we restrict attention scores to be computed only between nodes connected by at least one path of the flows.
Specifically, we employ a masking strategy~\citep{yan2021cate, luo2023transformers} with a mask matrix $M\in\mathbb{R}^{N\times N}$ defined as follows (refer to \cref{fig:attention} for an example of $M$): 
% that limits the scope of global attention to only focus on nodes that are connected by computational paths. Formally, a mask matrix is defined as, 
$$ M_{ij} = \begin{cases}
       1 & \text{if $v_i$ lies on any directed path from $v_j$}\\ 
       & \text{or $v_j$ lies on any directed path from $v_i$,}\\
       0 & \text{otherwise.}
   \end{cases} $$

%The primary goal of the global attention module is to capture features on a global scale. 
%In our flow-aware model, it is more crucial to focus attention on nodes that are directly connected by paths, rather than considering all node pairs, which reflect the flow through the neural architecture. 
%However, rather than considering all node pairs, our flow-aware model learns attention for nodes that are directly connected by paths, which reflect the flows through the neural architecture.  % SY: 수정안

Specifically, given the input node-embedding matrix $H^{(\ell-1)}\in\mathbb{R}^{N\times d}$ and the mask matrix $M$, the flow-aware global attention module computes its output node-embedding matrix $H^{(\ell)}_{global}\in\mathbb{R}^{N\times d}$ as follows:
\begin{equation}
    H^{(\ell)}_{global} = \operatorname{MMHA}(H^{(\ell-1)},H^{(\ell-1)},H^{(\ell-1)},M).~\label{eq:global}
\end{equation}
Here, $\operatorname{MMHA}$ is the Masked Multi-Head Attention module:
\begin{equation*}
    \operatorname{MMHA}(Q, K, V, M)=\operatorname{Concat}(\operatorname{head}_1, \dots, \operatorname{head}_s)W^0,~\label{eq:mmha}
\end{equation*}
where $W^0\in \mathbb{R}^{sd_v\times d}$ is the learnable projection matrix, $s$ is the number of heads, and
\begin{align*}
    &\operatorname{head}_i = \operatorname{Attn}(QW_i^Q, KW_i^K, VW_i^V, M),\\
    &\operatorname{Attn}(Q, K, V, M)=\left(M\odot \operatorname{Softmax}\left(\frac{QK^T}{\sqrt{d_k}}\right)\right)V.
\end{align*}
Here, $\odot$ is element-wise multiplication; and $W_i^Q \in \mathbb{R}^{d\times d_k}$, $W_i^K\in \mathbb{R}^{d\times d_k}$, and $W_i^V\in \mathbb{R}^{d\times d_v}$ denote $i$-th head's learnable query, key, and value projection matrices, respectively. We adhere to the condition $d_k=d_v=d/s$ for every head.

\subsection{Overall framework: \textbf{\method}} \label{method:overall}

The overall framework of \method is illustrated in \cref{fig:overview}.
% Suppose we have $L$ \flayer layers in total.
For each $\ell$, it derives the output node embedding matrix $H^{(\ell)}$ from  $H^{(\ell)}_{flow}$ (\cref{eq:flow}) and $H^{(\ell)}_{global}$ (\cref{eq:global}) for the $\ell$-th layer as follows:
\begin{equation}\label{eq:H_ell}
    H^{(\ell)} = \operatorname{FeedForward}(H^{(\ell)}_{flow} + H^{(\ell)}_{global})
\end{equation}
In our implementation, we employ a 2-layer MLP with ReLU activation~\cite{agarap2018deep} as the feedforward network.
As shown in \cref{fig:overview}, note that we incorporate skip-connection and batch normalization in every module.

The output $H^{(\ell)}$ is used as the input of the next \flayer layer, and for the first layer, we utilize a projected input feature matrix as the input by multiplying $X$ with a learnable projection matrix $P\in \mathbb{R}^{D \times d}$, i.e., $H^{(0)}=XP$.
% Note that 
Each \flayer layer has a separate set of learnable parameters.

%Suppose we have $L$ \flayer layers in total.
The node embeddings in the output $H^{(L)}$, where $L$ represents the total number of \flayer layers, are aggregated to drive the final embedding $z_{G}$ of the input neural-architecture graph $G$ as follows:
\begin{equation*}
    z_G = \operatorname{READOUT}(H^{(L)}), ~\label{eq:overall:readout}
\end{equation*}
For aggregation, we use mean pooling as the readout function in our implementation.

\smallsection{Application to performance prediction}
The architecture embedding $z_G$ is used for downstream tasks. For example, for performance prediction, it may serve as input to a regressor that outputs the estimated performance $\hat{y}_G$ as follows:
\begin{equation*}
    \hat{y}_G = \operatorname{Regressor}(z_G).~\label{eq:overall:predict}
\end{equation*}
In \cref{sec:experiments}, we employ a fully connected layer as the regressor and utilize the following margin ranking loss for training both \method and the regressor:
\begin{equation}
\mathcal{L} = \sum_{(i,j): y_i>y_j}\max(0, \text{margin} - (\hat{y}_i - \hat{y}_j)),
\end{equation}
where $y_i$ and $y_j$ are the ground-truth performances of architectures $G_i$ and $G_j$, respectively.
For each pair of architectures $G_i$ and $G_j$ in the training set such that $G_i$ outperforms $G_j$ (i.e., $y_i>y_j$), the loss encourages $\hat{y}_i$ to be greater than $\hat{y}_j$ by at least a specified margin.
Such designs for loss functions are commonly employed when it is important to make relative comparisons among instances (in our case, we compare neural architectures to recommend better ones).

% With the final layer output, $H^{(L)}$, we integrate all node embeddings using a readout function to derive a single embedding for the architecture. This yields the network embedding \( z_G\in\mathbb{R}^{d} \). We use global mean pooling for the readout function. To estimate performance, represented as \( \hat{y} \), we employ a fully connected layer(FC) as a regressor subsequent to the readout.
% \begin{align}
%     z_G &= \text{READOUT}(H^{(L)})~\label{eq:overall:readout}\\
%     \hat{y} &= \text{Regressor}(z_G)~\label{eq:overall:predict}
% \end{align}

%If the predicted rankings are correct and exceed the margin, the loss is zero; otherwise, the model incurs a penalty to encourage proper ranking.:
%The loss is computed as:
%where \( \delta_{ij} \) is set to 1 if the ground truth performance indicates that \( \hat{y}_i \) should indeed be higher than \( \hat{y}_j \), and it is set to -1 if the reverse is true. 

\section{Experiments}
\label{sec:experiments}

In this section, we review our experiments. For evaluation, we focus on the downstream task of predicting the performance of neural architectures.
In \cref{exp:overall}, we compare the accuracies of \method and six baseline methods, including two state-of-the-art methods (spec., TA-GATES~\cite{ning2022ta} and NAR-Former~\cite{yi2023nar}) using three performance prediction benchmark datasets composed of computer vision model architectures.
In \cref{exp:abl}, we conduct an ablation study to validate each component of \method.
In \cref{exp:extend}, we extend our evaluation to datasets consisting of graph neural networks and auto speech recognition models.
In \cref{exp:speed}, we examine the training and inference speed of \method.

\subsection{Experimental settings}
Below, we provide an overview of our experimental setup.

\subsubsection{Datasets}
We evaluate the effectiveness of neural architecture encoding methods using five benchmark datasets designed for performance prediction, spanning three domains:
\begin{itemize}
\item \textbf{Computer vision:} We use three datasets: NAS-Bench-101~\citep{ying2019bench, zela2019bench}, NAS-Bench-201~\citep{dong2020nasbench201}, and NAS-Bench-301 \citep{zela2022surrogate}. These datasets contain computer vision models.
\item \textbf{Speech recognition:} We employ NAS-Bench-ASR~\citep{mehrotra2020bench}, which consists of auto speech recognition architectures. 
\item \textbf{Graph learning:} We include NAS-Bench-Graph~\citep{qin2022bench}, which consists of graph neural networks.
\end{itemize}
Refer to \cref{tab:data}, for basic statistics, and the supplementary material,
for details including our preprocessing methods.

\subsubsection{Baseline methods}
%In our comparative analysis, \method is benchmarked against prominent models in graph representation learning,
We utilize six baseline approaches, categorized as follows:
 \textbf{(a) Graph neural networks:} GatedGCN~\citep{bresson2017residual} and directed acyclic graph neural network (DAGNN)~\citep{thost2021directed},  \textbf{(b) Graph transformers:} GraphGPS~\citep{rampavsek2022recipe} and DAGFormer~\citep{luo2023transformers}, and 
 \textbf{(c) Neural architecture encoders:} TA-GATES~\citep{ning2022ta} and NAR-Former~\citep{yi2023nar}, which are state-of-the-art methods for neural architecture performance prediction.
We use the official implementations of these methods, and the links can be found in the supplementary material.
%For our baseline methods, we employ graph neural networks, graph transformers, and two specializsed neural architecture encoding methods.
%For graph neural networks, we utilize  Gr. 
%For graph transformers, we adopt  
%For neural architecture encoding methods, we utilize , which are SOTA methods 
%The reported baseline performances are obtained from their respective official implementations.

\subsubsection{Training and Evaluation protocol} \label{exp:protocol}
For model training and evaluation, we follow the setting in \citep{ning2022ta}, including their training and test splits.
We use a subset of the training split as the actual training set, varying the size of this subset: 1\%, 5\%, 10\%, and 50\% of the training split.
We use the first 40 architectures in the test split as a validation set for {hyperparameter tuning} and early stopping, the remaining ones in the split as a test set.
In each setting, we perform 9 trials using the three different splits and three different random seeds, and we report the mean and standard deviation across these trials.
As accuracy metrics, we use Kendall's Tau~\citep{sen1968estimates} to assess overall performance and Precision@K (which measures the proportion of correctly predicted top-K architectures among the true top-K outperforming architectures) for the performance of identifying the best architectures.
Note that these metrics are commonly employed in the field of neural architecture encoding~\citep{ning2020generic, ning2022ta, yi2023nar}.
%For each dataset, we split it into train and test sets. 
%We maintain a constant test set while systematically varying the training ratio within the training subset. 
%We employ a fixed training and test set, while sys

\begin{table}[t]
    \centering
    \caption{Basic information about the benchmark datasets we used. The sizes of the training and test splits used in \citep{ning2022ta} are reported. Refer to \cref{exp:protocol} for details about training and test splits.}
    \label{tab:data}
    \scalebox{0.85}{
    \begin{tabular}{l|c|r|r}
        \thickhline
        Dataset & Domain & \# trains & \# tests \\
        \hline
        NAS-Bench-101 &\multirow{3}{*}{Computer vision} &  7,290& 7,290\\
        NAS-Bench-201 &  & 7,813& 7,812\\ 
        NAS-Bench-301 & & 5,896& 51,072\\
        \hline
        NAS-Bench-ASR & Speech recognition & 4,121&4,121 \\
        \hline
        NAS-Bench-Graph & Graph learning & 13,103&13,103 \\
        \thickhline
    \end{tabular}
    }
\end{table}

\begin{table*}[hbt!]
\vspace{-2mm}
  \centering
  \caption{Kendall's Tau (scaled up by a factor of 100, mean and standard deviation over 9 trials) on three datasets: NAS-Bench-101, NAS-Bench-201, NAS-Bench 301. In each setting, the best performances are highlighted in \green{green}. 
  \textbf{NA}: there is no trivial extension of \NARFo to NAS-Bench-301, which consists of two-cell architectures.
  Note that, in every setting, \method performs best. 
  }
  % \vspace{-1mm}
  \scalebox{0.75}{%
  \renewcommand{\arraystretch}{1.1}
    %\begin{tabular}{l|c|c|c|c|c|c|c|c}
    \begin{tabular}{l|c c c c|c c c c|c c c c|c}
        \thickhline
        Datasets& \multicolumn{4}{c|}{NAS-Bench-101}& \multicolumn{4}{c|}{NAS-Bench-201} & \multicolumn{4}{c|}{NAS-Bench-301}&Avg.\\
        %\midrule
        %\cmidrule{1-13}
        \cline{1-13}
        Training portions& 1\% & 5\% & 10\% & 50\%& 1\% & 5\% & 10\% & 50\%& 1\% & 5\% & 10\% & 50\%&Rank\\
        \hline\hline
        \GCN~\cite{bresson2017residual}
        & 67.4~\std{(6.0)}& 79.6~\std{(4.1)}& 82.0~\std{(5.1)} & 84.8~\std{(5.9)}
        & 70.9~\std{(1.8)}& 84.1~\std{(0.6)}& 88.6~\std{(0.3)} & 92.3~\std{(0.1)}
        & \secb 61.8~\std{(2.4)}& 70.0~\std{(0.9)}& 71.4~\std{(1.0)} & 72.7~\std{(1.5)}&4.91\\
        \DAGNN~\cite{thost2021directed}
        &72.4~\std{(4.5)}& \secb 82.9~\std{(3.1)}& \secb 84.4~\std{(4.4)} & 85.9~\std{(5.3)}
        & 75.8~\std{(1.0)}& \secb 87.5~\std{(0.8)}& \secb 90.6~\std{(0.2)} & 92.6~\std{(0.0)}
        & 61.5~\std{(1.9)}& \secb 70.9~\std{(0.5)}& \secb 73.4~\std{(1.2)} & \secb 76.1~\std{(1.3)}&2.50\\
        \GPS~\cite{rampavsek2022recipe}
        & 70.6~\std{(4.4)}& 81.7~\std{(3.8)}& 83.9~\std{(4.2)} & 85.9~\std{(5.1)}
        & 71.3~\std{(1.3)}& 82.5~\std{(0.6)}&87.8~\std{(0.5)} & \secb 92.7~\std{(0.1)}
        & 59.7~\std{(1.8)}& 69.3~\std{(0.9)}& 70.7~\std{(1.2)} & 73.8~\std{(0.7)}&4.75\\
        \DAGFo~\cite{luo2023transformers}
        & \secb 73.0~\std{(4.3)} & 75.6~\std{(5.2)}& 77.2~\std{(7.0)} & 80.9~\std{(5.9)}
        & 73.0~\std{(73.0)} & 84.9~\std{(0.8)}&88.8~\std{(0.5)} &92.7~\std{(0.1)}
        & 61.3~\std{(2.0)}& 70.7~\std{(0.8)}& 72.1~\std{(0.8)} & 74.8~\std{(1.0)} &3.91\\
        \NARFo~\cite{yi2023nar}
        & 59.4~\std{(8.8)}& 72.0~\std{(8.2)}& 75.5~\std{(10.2)} & 79.8~\std{(5.9)}
        & 62.3~\std{(4.0)}& 80.7~\std{(1.8)}& 87.3~\std{(0.7)} & 88.9~\std{(0.3)}
        & \textbf{NA}& \textbf{NA}& \textbf{NA} & \textbf{NA} &\textbf{-}\\
        \TAGATES~\cite{ning2022ta}
        & 70.8~\std{(6.0)}& 82.3~\std{(2.7)}& 83.9~\std{(3.5)} & \secb 86.3~\std{(3.9)}
        & \secb 77.7~\std{(1.7)}& 86.3~\std{(0.8)}& 88.7~\std{(0.3)} & 91.4~\std{(0.5)}
        & 61.3~\std{(1.2)}& 68.9~\std{(1.6)}& 71.8~\std{(1.6)} &75.4~\std{(0.7)} &3.83\\
        \hline\hline
        \method
        & \best 75.0~\std{(2.9)} & \best 86.1~\std{(0.8)} & \best88.1~\std{(0.2)} & \best89.6~\std{(0.1)}
        & \best80.0~\std{(0.8)} & \best89.8~\std{(0.3)} & \best91.3~\std{(0.2)} & \best92.9~\std{(0.1)}
        & \best 64.2~\std{(1.6)} & \best 72.2~\std{(1.0)} & \best 73.6~\std{(1.3)} & \best 77.5~\std{(0.7)} & \best 1.00\\
    \thickhline
    \end{tabular}
    }
    \label{tab:main}
\end{table*}

\begin{table*}[hbt!]
\vspace{-2mm}
  \centering
  \caption{Precision@K (scaled up by a factor of 100, mean and standard deviation of 9 trials). The proportion of training samples is fixed to 5\%. In each setting, the best performances are highlighted in \green{green}. 
  \textbf{NA}: there is no trivial extension of \NARFo to NAS-Bench-301, which consists of two-cell architectures.
  Note that, in most cases, \method identifies top-k architectures most accurately. 
  }
  % \vspace{-1mm}
  
  \scalebox{0.75}{%
  \renewcommand{\arraystretch}{1.1}
    %\begin{tabular}{l|c|c|c|c|c|c|c|c}
    \begin{tabular}{l|c c c c|c c c c|c c c c|c}
        \thickhline
        Datasets& \multicolumn{4}{c|}{NAS-Bench-101 (5\%)}& \multicolumn{4}{c|}{NAS-Bench-201 (5\%)} & \multicolumn{4}{c|}{NAS-Bench-301 (5\%)} & Avg.\\
        \cline{1-13}
        K (for P@Top K\%) & 1 & 5 & 10 & 50 & 1 & 5 & 10 & 50 & 1 & 5 & 10 & 50 & Rank\\
        \hline\hline
        \GCN~\cite{bresson2017residual}
        &44.4~\std{(7.6)}&65.6~\std{(3.5)}&76.2~\std{(2.7)}&90.5~\std{(2.1)}
        &42.3~\std{(3.7)}&68.5~\std{(3.1)}&80.9~\std{(1.9)}&94.1~\std{(0.6)}
        &19.1~\std{(4.1)}&55.2~\std{(4.5)}&71.8~\std{(2.9)}&85.4~\std{(0.4)} & 4.83\\
        \DAGNN~\cite{thost2021directed}
        &41.7~\std{(5.9)}&65.4~\std{(4.2)}&\secb 79.3~\std{(2.9)}&\secb 92.0~\std{(1.3)}
        &49.6~\std{(6.2)}&69.7~\std{(3.0)}&\secb 83.1~\std{(0.7)}&\secb 95.3~\std{(0.9)}
        &\best 23.1~\std{(2.1)}&\secb 58.3~\std{(3.4)}& 73.1~\std{(1.5)} & \secb 85.8~\std{(0.4)} & 2.75\\
        \GPS~\cite{rampavsek2022recipe}
        &44.3~\std{(12.2)}&\secb 67.1~\std{(2.7)}&78.7~\std{(1.9)}&91.2~\std{(2.0)}
        & 49.4~\std{(4.6)}& 67.9~\std{(4.9)}& 78.9~\std{(3.4)} & 93.4~\std{(0.3)}
        &20.6~\std{(2.1)}&57.2~\std{(3.8)}&73.4~\std{(2.5)}&84.8~\std{(0.5)} & 4.17\\
        \DAGFo~\cite{luo2023transformers}
        &39.4~\std{(7.9)}&61.8~\std{(5.6)}&71.6~\std{(5.0)}&88.2~\std{(2.4)}
        &\secb 50.7~\std{(5.8)} &\secb 70.4~\std{(2.9)} & 82.5~\std{(2.3)}& 94.2~\std{(0.5)}
        &20.7~\std{(3.4)}&57.6~\std{(3.7)}&\secb 73.4~\std{(2.5)}&85.6~\std{(0.4)} & 3.83\\
        \NARFo~\cite{yi2023nar}
        & \best 47.2~\std{(9.9)}& 62.6~\std{(7.9)}& 67.8~\std{(8.4)} & 85.9~\std{(5.2)}
        &49.5~\std{(6.5)}& 64.7~\std{(2.0)}& 69.9~\std{(2.0)} & 92.3~\std{(1.0)}
        & \textbf{NA}& \textbf{NA}& \textbf{NA} & \textbf{NA} & \textbf{-}\\
        \TAGATES~\cite{ning2022ta}
        &44.6~\std{(9.7)}&66.6~\std{(4.0)}&78.1~\std{(4.6)}&91.8~\std{(1.2)}
        &49.4~\std{(3.1)}&66.7~\std{(3.3)}&78.1~\std{(2.8)}&94.8~\std{(0.7)}
       &20.1~\std{(5.0)}&56.2~\std{(6.1)}&72.4~\std{(3.6)}&84.7~\std{(0.7)} & 4.33\\
       \hline\hline
        \method
        &\secb46.5~\std{(11.2)}&\best70.0~\std{(1.5)}&\best80.9~\std{(1.8)}&\best92.7~\std{(1.7)}
        &\best57.0~\std{(5.4)}&\best74.7~\std{(1.8)}&\best85.2~\std{(1.3)}&\best96.9~\std{(0.7)}
        &\secb20.8~\std{(3.7)}&\best58.5~\std{(2.5)}&\best74.7~\std{(2.4)}&\best86.6~\std{(0.6)} &\best  1.08\\
    \thickhline
    \end{tabular}
    }
    \label{tab:pak}
\end{table*}

\subsection{Performance on computer vision benchmarks} \label{exp:overall}

In this subsection, we focus on the computer vision benchmarks for which we have extensive baseline methods.
In \cref{tab:main,tab:pak}, we report the performance prediction accuracies of the considered methods using two metrics across different training instance ratios.
Notably, \method consistently outperforms all baseline methods across all settings in terms of Kendall's Tau. In terms of Precision@K, it performs best in 10 out of 12 settings, ranking second in the other settings. Two key observations are as follows.

%Notably, under Kendall's Tau evaluation metric, \method outperforms all the used baseline methods in all the settings.
%Moreover, under the Precision@K evaluation metric, \method outperforms all the used baseline methods in 10 out of 12 settings and shows the second-best performance in the other two. 
The suboptimal performance of GraphGPS indicates that a graph transformer alone is insufficient in effectively representing neural architectures. 
Specifically, in terms of Kendall's Tau, the performance gap can be as large as 8.7 percentage points between \method and GPS. 
We, thus, argue that our incorporation of information flows into a graph transformer,  through the introduction of the flow encode module and the flow-aware global attention module, plays a pivotal role in \method's success.

The superiority of \method over TA-GATES highlights the importance of the global attention mechanism. While TA-GATES may capture the information flow at a local-level through its information propagation scheme, it does not adequately leverage the global context of neural architectures. 
\method, on the other hand, uses the global attention mechanism to capture the graph-level (i.e., architecture-level) characteristics, empowering \method to yield better representations of architectures.
%This advantage encourages \method to more accurately represent the input architecture.

In summary, our empirical findings substantiate that \method serves as an effective predictor of neural architecture performance. % in the computer vision domain. Refer to \cref{exp:extend} for results in other domains.

\begin{table}[t!]
  \centering
  \caption{
    Comparison with four variants of \method in terms of Kendall's Tau, using the same setups as in \cref{tab:main}. In each setting, the best performances are highlighted in \green{green}. \textbf{AS}: Asynchronous message passing. \textbf{FB}: Forward-backward message passing. \textbf{GA}: Global attention. In most cases, \method, which is equipped with all components, outperforms all of its variants, thereby validating the effectiveness of each component.}
  % \vspace{-1mm}
  \scalebox{0.7}{%
  
    %\begin{tabular}{l|c|c|c|c|c|c|c|c}
    \begin{tabular}{l|c|ccc|c c c c}
        \thickhline
        \multirow{2}{*}{Dataset} &\multirow{2}{*}{\#}& \multicolumn{3}{c|}{Components}& \multicolumn{4}{c}{Training Portions} \\
        \cline{3-9}
         && AS & FB & GA& 1\% & 5\% & 10\% & 50\%   \\
        \hline\hline
        \multirow{5}{*}{NB 101} & \textbf{(1)}
        &\no&\no& \yes 
        & 41.5\std{(1.7)}& 42.5~\std{(1.6)}& 41.1~\std{(2.8)}& 43.1~\std{(1.4)}\\
        & \textbf{(2)}&\no&\yes& \yes
        & 65.5~\std{(8.8)}& 56.8~\std{(5.0)}&53.8~\std{(7.4)} & 69.6~\std{(11.6)}\\
         &\textbf{(3)}&\yes&\no& \yes 
        & \best 76.7~\std{(4.1)}&\secb 83.9~\std{(2.6)}& \secb 84.6~\std{(4.0} &  \secb 85.6~\std{(5.4)}\\
        & \textbf{(4)}&\yes&\yes &\no
        & \secb 76.5~\std{(3.0)}& 83.2~\std{(3.9)}&83.9~\std{(5.1)} & 85.3~\std{(6.3)} \\   
        &\textbf{-}& \yes&\yes&\yes 
         & 75.0~\std{(2.9)} & \best 86.1~\std{(0.8)} &\best 88.1~\std{(0.2)} & \best89.6~\std{(0.1)} \\
        \hline
        \multirow{5}{*}{NB 201}
        &\textbf{(1)}&\no&\no& \yes 
        & 75.9~\std{(1.2)}& 86.5~\std{(0.2}& 88.2~\std{(0.3)}
        &  89.7~\std{(0.1)}\\
        & \textbf{(2)}&\no&\yes& \yes
        & 73.7~\std{(1.1)}&85.6~\std{(0.7)} & 89.2~\std{(0.5)}
        &  92.9~\std{(0.1)}\\
        &\textbf{(3)}&\yes&\no& \yes 
        & 76.2~\std{(2.1)}& 88.6~\std{(0.8)}& 90.9~\std{(0.1)}
        & \secb92.9~\std{(0.1)} \\
        & \textbf{(4)} &\yes&\yes &\no
        & \best79.3~\std{(1.2)}&\secb 89.5~\std{(0.5)} & \secb 91.1~\std{(0.3)}
        &  92.9~\std{(0.3)}\\
        & \textbf{-}&\yes&\yes&\yes
         & \secb 79.0~\std{(0.8)} & \best89.8~\std{(0.3)} & \best91.3~\std{(0.2)} & \best92.9~\std{(0.1)}\\
        \hline
        \multirow{4}{*}{NB 301} 
        & \textbf{(1)}&\no&\no& \yes
        & \secb 63.3\std{(2.7)}& 69.0~\std{(2.8)}& 68.1~\std{(2.9)}
        & 59.8~\std{(2.3)} \\
        & \textbf{(3)} &\yes&\no& \yes
         & 59.5\std{(2.6)}& 69.0~\std{(1.8)}& 50.2~\std{(15.0)}
        & 45.3~\std{(19.4)} \\
        & \textbf{(4)}&\yes&\yes &\no
        & 60.9~\std{(3.1)}& \secb 69.8~\std{(1.4)}& \secb 70.9~\std{(1.2)}
        & \secb 67.7~\std{(3.1)} \\
        & \textbf{-}&\yes&\yes&\yes
         & \best 64.2~\std{(1.6)} & \best 72.2~\std{(1.0)} & \best 73.6~\std{(1.3)} & \best77.5~\std{(0.7)} \\
    \thickhline
    \end{tabular}
    }
    \vspace{3mm}
    \label{tab:abl}
\end{table}

\begin{table}[t!]
\vspace{-2mm}
  \centering
  \caption{Kendall's Tau (scaled up by a factor of 100, mean and standard deviation of 9 experiments) on two datasets beyond the computer vision domain: NAS-Bench-Graph (NB-G)~\citep{qin2022bench} and  NAS-Bench-ASR (NB-ASR)~\citep{mehrotra2020bench}. In each setting, the best performances are highlighted in \green{green}. In most cases, \method performs best.
  }
  \label{tab:induct}
  % \vspace{-1mm}
  \scalebox{0.75}{%
    %\begin{tabular}{l|c|c|c|c|c|c|c|c}
    \begin{tabular}{c|l|c c c c}
        \thickhline
        \multirow{2}{*}{Dataset}& \multirow{2}{*}{Encoder}& \multicolumn{4}{c}{Training portions}   \\
        \cline{3-6}
        & & 1\% & 5\% & 10\% & 50\% \\
        \hline\hline
        \multirow{4}{*}{NB-G}& \DAGNN~\cite{thost2021directed}
        & \secb 48.1~\std{(3.2)}& \secb 64.4~\std{(1.2)}& \secb 67.4~\std{(1.1)} & \best 73.1~\std{(0.8)}\\
        & \DAGFo~\cite{luo2023transformers}
        & 47.9~\std{(0.6)}& 60.8~\std{(1.6)}& 64.9~\std{(1.0)} & {72.4}~\std{(0.3)}\\
        & \TAGATES~\cite{ning2022ta}
        & 33.1~\std{(1.4)}& 34.1~\std{(2.0)}& 35.4~\std{(0.8)} & 35.7~\std{(0.5)}\\
        \cline{2-6}
         &\method
         & \best 49.5~\std{(1.1)} & \best 65.9~\std{(1.3)} & \best 68.9~\std{(0.6)} & \secb 72.7~\std{(0.2)} \\
                \hline
        \multirow{4}{*}{NB-ASR}& \DAGNN~\cite{thost2021directed}
        & 29.5~\std{(3.9)}& {40.9}~\std{(2.4)}& {45.2}~\std{(1.3)} & 44.0~\std{(0.4)}\\
        & \DAGFo~\cite{luo2023transformers}
        &  29.9~\std{(5.4)}& \secb 42.5~\std{(1.1)}& \secb 45.3~\std{(1.0)} & {34.6}~\std{(5.8)}\\
        & \TAGATES~\cite{ning2022ta}
        &\best 34.0~\std{(2.3)} & 41.4~\std{(2.0)}& 44.9~\std{(2.2)}&\secb 50.9~\std{(0.8)}\\
        \cline{2-6}
         &\method
         & \secb 31.1~\std{(8.0)} & \best44.0~\std{(0.9)} & \best 47.3~\std{(1.3)} & \best 52.2~\std{(1.4)} \\
    \thickhline
    \end{tabular}
    }
\end{table}

\subsection{Ablation studies} \label{exp:abl}

In this subsection, we conduct ablation studies to validate the design choices made in \method. Specifically, we aim to analyze the necessity of (a) asynchronous message-passing, (b) forward-backward message-passing, and (c) flow-aware global attention.
To this end, we use four variants of \method \textbf{(1)} without the flow encode module (i.e., eliminating both asynchronous and forward-backward message passing), \textbf{(2)} without asynchronous message passing, \textbf{(3)} without forward-backward message passing, and \textbf{(4)} without flow-aware global attention. 

As shown in \cref{tab:abl}, \method, which is equipped with all the components, consistently outperforms all variants in most settings, confirming the efficacy of our design choices. Further observations deserve attention. First, the necessity of asynchronous message passing for capturing flows is confirmed by the superior performance of \textbf{(3)} over \textbf{(1)}, and that of \method over \textbf{(2)}. 
Second, the advantage of forward-backward message passing is demonstrated by \method's superiority over \textbf{(3)}.
Lastly, incorporating flow-awareness into global attention is advantageous, as evidenced by \method's advantage over variant \textbf{(4)}.

\subsection{Performance in various domains} \label{exp:extend}
Since our input modeling does not require complex preprocessing, it can be readily applied to architectures across various domains. 
We apply \method to graph neural networks on NAS-Bench-Graph and automatic speech recognition architectures on  NAS-Bench-ASR. Among the baseline methods used in \cref{exp:overall}, we use the best method of each type: DAGNN, DAGFormer, and TA-GATES.

As shown in \cref{tab:induct}, \method consistently performs best in most cases. 
These results indicate that \method effectively captures important architectural characteristics across various domains. %, underscoring its adaptability regardless of the architecture's domain. 
\TAGATES, which is tailored for encoding architectures in the computer vision domain, also shows strong performance in the domain of automatic speech recognition. \TAGATES effectively updates operation embeddings by multiplying operation embeddings and input information vectors, which is akin to convolutional mechanisms prevalent in auto speech recognition architectures. However, its effectiveness diminishes in scenarios where message passing between nodes, a key characteristic of graph neural networks, is required.

\subsection{Training and inference speed} \label{exp:speed}
%In this section, we outline the training and inference speeds of \method as detailed in~\cref{tab:speed}. 
In this subsection, we compare the training and inference speeds of \method and two state-of-the-art neural architecture encoding methods: \NARFo and \TAGATES.
To this end, we train all the models for 200 epochs with a batch size of 128, using an NVIDIA RTX 2080 GPU.
We use NAS-Bench-101 with a training ratio of 1\%.
For a fair comparison, we exclude all additional time-consuming training strategies of \NARFo and \TAGATES (e.g., input augmentation) in this experiment.
%as detailed in~\cref{tab:speed}. 
%The table compares the training and inference times of our model with neural architecture encoding models. 
%These measurements are based on a batch size of 128, over 200 learning epochs, with a training ratio of 1\% using the NAS-Bench-101 dataset. 
%All tests were carried out under uniform conditions on identical hardware, utilizing an NVIDIA RTX 2080 GPU. 
%For a fair comparison, we \red{do not include the augmentation} of \NARFo for this experiment.

As shown in \cref{tab:speed}, \method takes the shortest training time among the three methods.
In particular, training \method is $4.44\times$ faster than training \NARFo. 
This substantial speed advantage stems from the notable difference in model sizes, with \NARFo having $5.35\times$ the number of parameters compared to \method.
Compared to \TAGATES, \method exhibits a slight speed advantage.
Despite the small model size of \TAGATES, our specialized batch operations boost the training of \method.
Refer to the supplementary material for details of the batch operations.
%\method, having 

In terms of inference time, there is not much difference among the three methods, and \method ranks second.
In practical scenarios, neural architecture performance prediction involves collecting labels (e.g., ground-truth performance) for the architectures in the training set, which requires time-consuming training of the architectures.
For example, in the case of NAS-Bench-101, training just 1\% of the architectures can take up to 24 GPU hours.
Thus, inference speed is not a bottleneck, due to the extensive computational cost of training.

\begin{table}[t]
    \vspace{-3mm}
    \centering
    \caption{Training and inference times on NAS-Bench-101 with a training ratio of 1\%, 200 epochs, and a batch size of 128.}
    \label{tab:speed}
    \scalebox{0.75}{
    \begin{tabular}{l|r|r|r}
        \thickhline
        Encoder & Training time (sec) & Inference time (sec) & \# Params \\ 
        \hline
        % \GCN & 14.26~\std{(0.29)} & 2.06~\std{(0.06)}\\
        % \DAGNN & 74.61~\std{(2.28)} & 2.95~\std{(0.03)}\\ 
        % \GPS & 14.70~\std{(0.56)} & 2.13~\std{(0.03)}\\
        % \DAGFo &16.90~\std{(0.57)} & 2.12~\std{(0.03)} \\
        \NARFo~\cite{yi2023nar} &278.13~\std{(13.01)} & 2.55~\std{(0.07)} & 4,882,081 \\
        \TAGATES~\cite{ning2022ta} &62.66~\std{(0.54)} & 3.01~\std{(0.21)} & 348,065\\
        \method &58.08~\std{(1.39)} & 2.94~\std{(0.07)} & 901,459\\
        \thickhline
    \end{tabular}
    }
\end{table}

\section{Conclusions}
\label{sec:conclusion}
In this work, we propose \method, a novel graph transformer model designed for neural architecture encoding.
\method excels at capturing information flows within neural architectures, considering both local and global aspects.
Through comprehensive evaluations across five benchmarks for architecture performance prediction,  \method exhibits significant and consistent superiority over several state-of-the-art baseline methods, ultimately achieving state-of-the-art performance.
Notably, \method's superiority extends beyond computer-vision architectures, demonstrating its effectiveness for graph-learning and speech-recognition architectures. 
%This versatility underlines \method's adaptability and its potential as a tool in various architectural domains.

{
\small 
\section*{Acknowledgements}
This work was supported by Institute of Information \& Communications Technology Planning \& Evaluation (IITP) grant funded by the Korea government (MSIT) (No. 2022-0-00871, Development of AI Autonomy and Knowledge Enhancement for AI Agent Collaboration) (No. 2019-0-00075, Artificial Intelligence Graduate School Program (KAIST)).
}

{
    \small
    \bibliographystyle{ieeenat_fullname}
    \bibliography{bib}
}

\newpage
\appendix
\section{Dataset description} %explanation

In this section, we provide detailed descriptions of the datasets used.

\begin{itemize}
\item \textbf{NAS-Bench-101~\citep{ying2019bench}}
is a dataset with 423K architectures trained on the CIFAR-10~\citep{krizhevsky2009learning} dataset. 
NAS-Bench-101 has an operation-on-node (OON) search space~\cite{ning2020generic, ning2022ta}. 
Following~\citet{ning2022ta}, we used the same subset of the NAS-Bench-101 dataset.
This subset consists of 14,580 architectures.

\item \textbf{NAS-Bench-201~\citep{dong2020nasbench201}} 
is a dataset with 15K architectures trained on the CIFAR-10 dataset.
We transformed the dataset, which is originally operation-on-edge (OOE)-based, into the OON format.

\item \textbf{NAS-Bench-301~\citep{zela2022surrogate}}
is a surrogate benchmark with 57K architectures each of which consists of two cells (spec., normal and reduction cells).
This dataset is originally OOE-based, and we converted the dataset into the OON format.
Following~\citet{ning2022ta}, we only used the anchor architecture-performance pairs.

\item \textbf{NAS-Bench-ASR~\citep{mehrotra2020bench}}
is a dataset with 8K architectures of auto speech recognition models, trained on the TIMIT audio dataset~\citep{garofolo1993darpa}.
We transformed the dataset, which is originally OOE-based, into the OON format.

\item \textbf{NAS-Bench-Graph~\citep{qin2022bench}}
is a dataset with 26K architectures of graph neural networks, trained on the Cora dataset~\citep{sen2008collective}.
Since the dataset is originally OON-based, no additional transformation is required.
\end{itemize}

\section{Experimental Details}
\subsection{Implementation details}
In this subsection, we provide several implementation details of \method.

\smallsection{Code implementation} 
We employed the framework of \GPS~\cite{rampavsek2022recipe} as the backbone to implement \method with Python 3.10, Pytorch 1.13.1, and Pytorch Geometric 2.2.0.

\smallsection{Obtaining representations in two-cell datasets} 
To obtain representations in two-cell-based datasets (e.g., NAS-Bench-301), we used the following projection strategy:

Let $h_{1, o} \in \mathbb{R}^{d}$ and $h_{2, o}\in \mathbb{R}^{d}$ denote the embeddings of the output nodes of cell 1 and cell 2 after forward message passing. 
Then, we concatenated $h_{1, o}$ and $h_{2, o}$ and projected the concatenated embeddings with a learnable projection matrix $W^{P} \in \mathbb{R}^{2d \times 2d}$, as follows:
\begin{equation}\label{eq:updated_embs}
   h' = \operatorname{concat}{(h_{1, o}, h_{2, o})} W^\text{P}. 
\end{equation}
Then, we split $h'\in\mathbb{R}^{2d}$ into two and regarded each split  as an embedding of the output nodes:
\begin{align}
    h_{1, o} &= (h'_1, h'_2, \cdots, h'_d), \\ 
    h_{2, o} &= (h'_{d+1}, h'_{d+2}, \cdots, h'_{2d}),
\end{align}
where $h'_i$ is the $i$-th entry of $h'$. Finally, we started asynchronous backward message passing (step 3 in Algorithm 1 of the main paper) with updated $h_{1,o}$ and $h_{2,o}$.

\smallsection{Training and hyperparameters} 
We used the AdamW optimizer~\citep{loshchilov2017decoupled} to train the models, and the best parameters were selected using early stopping. The hyperparameter search space was as follows:
\begin{itemize}
    \item $\text{lr}\in [10^{-4}, 10^{-2}]$
    \item $\text{weight decay} \in [10^{-10}, 10^{-3}]$
    \item $\text{margin} \in \{ 0.01, 0.05, 0.1, 0.5, 1.0\}$
    \item $L  \in \{ 4, 5, 6, 7, 8, 9, 10 \}$
    \item $d  \in \{ 64, 128, 256, 512 \}$
    \item $s  \in \{ 4, 8 \}$
    \item $\text{dropout} \in \{ 0.1, 0.2, 0.3, 0.4, 0.5\} $
\end{itemize}
Further details regarding hyperparameters, including the best hyperparameter combination in each dataset, are available at \url{http://github.com/y0ngjaenius/CVPR2024_FLOWERFormer}.

\subsection{Batch operation}
Asynchronous message passing inevitably introduces some delay, since each operation should be performed in a sequential manner.
In order to accelerate the computation, we employed group-based batch processing. Specifically, we used the topological batching strategy~\citep{crouse2019improving, thost2021directed}, which is specialized in handling asynchronous operations.
First, we grouped nodes that belong to the same topological generation and regarded each group as a single batch.
Then, instead of updating a representation of a single node at a time, we updated representations of nodes that belong to the same batch simultaneously.
Note that this simultaneous update process ensures the same result as updating each node in a batch one by one since the updating processes of nodes in the same topological generation are independent of each other.
In this manner, for a one-way message passing, we performed $\vert \mathcal{T} ^G \vert$ operations, which is generally smaller than the number of nodes.

\subsection{Baseline methods}
\begin{itemize}
\item \textbf{\GCN~\citep{bresson2017residual} and \GPS~\citep{rampavsek2022recipe}:} 
We used the \GCN implementation provided by the \GPS repository. For \GPS, we used GatedGCN and Performer as the MPNN and attention modules, respectively. We followed the choice used for OGBG-CODE2, which is the only dataset modeled as a DAG in \cite{rampavsek2022recipe}. The GitHub repository is \url{https://github.com/rampasek/GraphGPS}

\item \textbf{\DAGNN~\citep{thost2021directed}:}
This model has a bidirectional option, and we considered whether to use it or not as a hyperparameter. The GitHub repository is \url{https://github.com/vthost/DAGNN}

\item \textbf{\DAGFo~\citep{luo2023transformers}:}
\DAGFo introduces a framework that is applicable to existing graph transformers, We used the DAG+GraphGPS setting, which uses depth positional encoding and replaces the attention module of GraphGPS with reachability attention. The GitHub repository is \url{https://github.com/LUOyk1999/DAGformer}

\item \textbf{\NARFo~\citep{yi2023nar}:}
We followed the augmentation technique and hyperparameter setting of \NARFo used in \citep{yi2023nar} for each dataset. The GitHub repository is \url{https://github.com/yuny220/NAR-Former}

\item \textbf{\TAGATES~\citep{ning2022ta}:}
We followed the hyperparameter setting of \TAGATES used in \citep{ning2022ta} for each dataset. While the NAS-Bench-ASR dataset is OOE-based with multi-edges, the original TA-GATES implementation does not support multi-edges. Therefore, we converted the dataset into the OON format. The GitHub repository is \url{https://github.com/walkerning/aw_nas}
\end{itemize}

\section{Additional Experiments and Results}

\subsection{Neural architectures search experiments}
\begin{figure}[h]
    \centering
    \caption{The average test error of the best neural architectures obtained by the NPENAS algorithm using different performance predictors on the NAS-Bench-101 dataset over 100 trials. The plot shows that \method consistently outperforms other predictors in achieving lower test error rates, establishing its superiority in guiding the NAS process toward more accurate architectural choices.}
    \includegraphics[width=0.9\linewidth]{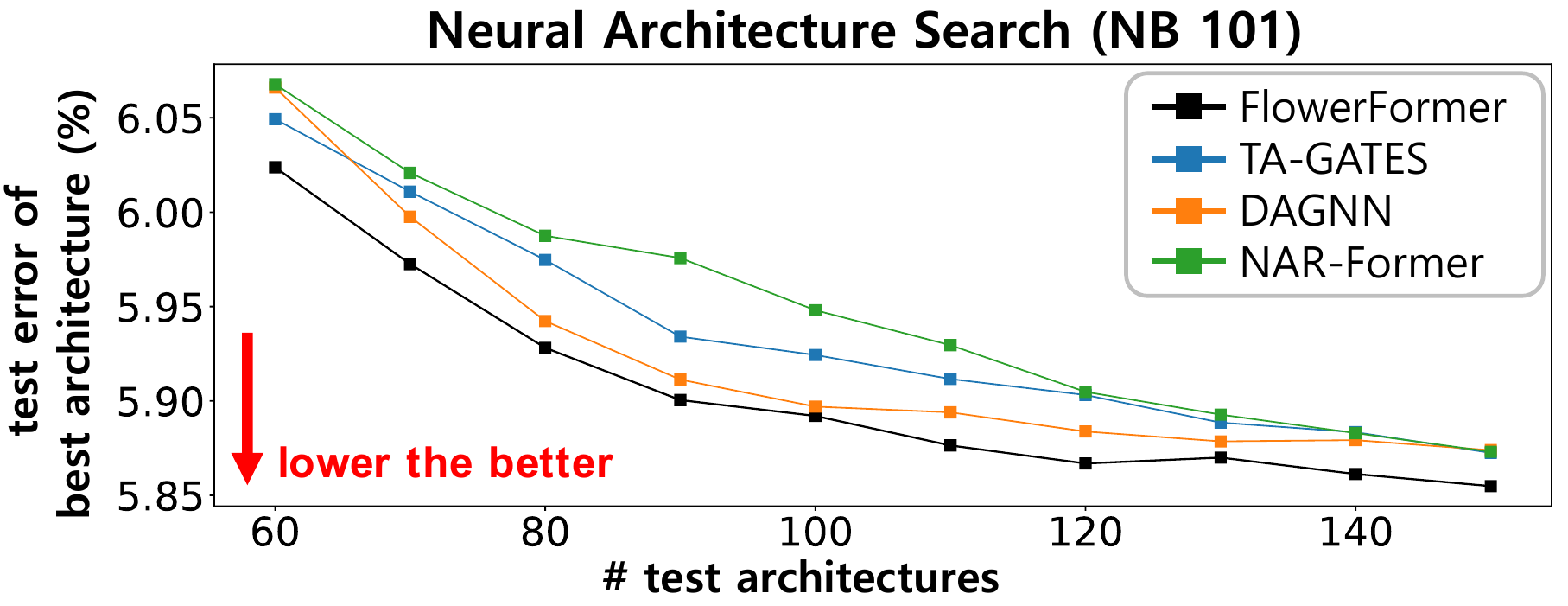} \\
    \label{nas}
\end{figure}

To validate the practical utility of \method, we conduct a series of Neural Architecture Search (NAS) experiments. We employ NPENAS~\cite{wei2022npenas2} as the backbone search algorithm, using TA-GATES, DAGNN, NAR-Former, and \method as performance predictors. 
We follow the experimental setup suggested in \citep{wei2022npenas2},with the modification of conducting 100 trials.
The results in Figure~\ref{nas} substantiate \method's superior performance compared to baseline methods.

\subsection{Latency prediction experiments}
To measure the encoding quality of \method in various aspects and validate its effectiveness, we conduct a latency prediction experiment on NAS-Bench-201, comparing \method with NAR-Former~\cite{yi2023nar}. 
For this comparison, we utilize Mean Absolute Percentage Error (MAPE) and Error Bound Accuracy (Acc($\delta$)), the same metrics employed by~\citet{yi2023nar} for latency prediction. 
As shown in Table~\ref{tab:mape_exps}, \method outperforms NAR-Former in the latency prediction task.

\begin{table}[h]
    \centering
    \caption{Mean Absolute Percentage Error (MAPE) and Error Bound Accuracy (ACC) at $\delta$ (scaled up by a factor of 100, mean over 9 trials) of latency prediction on the NAS-Bench-201 dataset. In each setting, the best performances are highlighted in \green{green}.\label{tab:mape_exps}}
    \scalebox{0.6}{
    \renewcommand{\arraystretch}{1.2} % Increased spacing for more room above and below
    \begin{tabular}{l|cc|cc|cc|cc}
        \thickhline
        Metric & \multicolumn{2}{c|}{MAPE$\downarrow$} & \multicolumn{2}{c|}{ACC ($\delta=0.1\%)\uparrow$} & \multicolumn{2}{c|}{ACC ($\delta=1\%)\uparrow$} & \multicolumn{2}{c}{ACC ($\delta=5\%)\uparrow$} \\
        \hline
        Training ratio & 5\% & 10\% & 5\% & 10\% & 5\% & 10\% & 5\% & 10\% \\
        \hline\hline % Double line for clear separation
        NAR-Former & 3.1 & 3.0 & 2.3 & 2.3 & 21.9 & 22.9 & 80.8 & 82.2 \\
        \method &  \best{1.1}& \best{0.9}&\best{8.6}&\best{12.7}&\best{67.2}&  \best{78.3}& \best{97.4}&\best{97.0}\\
        \thickhline
    \end{tabular}
    }
\end{table}

\subsection{Evaluation with additional metrics}
To evaluate the superior performance of \method across different evaluation criteria, we examine performance on NAS-Bench-101 using the Pearson Coefficient of Linear Correlation (LC) and Root Mean Squared Error (RMSE).
As shown in Table~\ref{tab:othermetrics}, \method shows the best performance in all the settings.

\begin{table}[h]
    \centering
    \caption{Linear Correlation (LC) and Root Mean Squared Error (RMSE) (mean over 9 trials) on the NAS-Bench-101 dataset. In each setting, the best performances are highlighted in \green{green}.\label{tab:othermetrics}}
    \scalebox{0.58}{
    \renewcommand{\arraystretch}{1.2}
    \begin{tabular}{l|cccc|cccc}
        \thickhline
         Metric & \multicolumn{4}{c|}{LC$\uparrow$} & \multicolumn{4}{c}{RMSE$\downarrow$} \\
        \hline
         Training ratio & 1\% & 5\% & 10\% & 50\% & 1\% & 5\% & 10\% & 50\% \\
        \hline
        \hline
         DAGNN & 0.4381 & 0.4919 & 0.5201 & 0.5876 & 0.0813 & 0.0802 & 0.0791 & 0.0755 \\
         TA-GATES & 0.3303 & 0.3432 & 0.5087 & 0.5677 & 0.0834 & 0.0831 & 0.0803 & 0.0777 \\
          \method & \best{0.5636} & \best{0.6583}& \best{0.6605} &	\best{0.7483} & \best{0.0768} & \best{0.0694} & \best{0.0670}	 &\best{0.0614}\\
        \thickhline
    \end{tabular}
    }
\end{table}
\subsection{Extended evaluation on additional dataset}
In this section, we analyze the performance of \method on ENAS, an additional dataset consisting of two cells. As shown in Table~\ref{tab:enas}, \method achieves the second-best performance in the dataset. 
%The primary challenge with the ENAS dataset stems from its two-cell nature, specifically involving normal and reduction cells. \method may face difficulties in datasets featuring a two-cell structure because the flow encode module, responsible for facilitating interaction between the two cells, is the sole component designed for this purpose.
We hypothesize that the sub-optimal performance of \method stems from its failure to account for interactions between two cells. Although there is information flow between two cells, \method lacks a dedicated global attention module that can capture their interactions.
This limitation suggests that enhancing the global attention module to incorporate strategies like cross-attention could be a valuable future research direction.

%It is noteworthy that the two-cell configuration of datasets does not universally hinder \method's performance, as evidenced by our success on the NB301 dataset, which also features a two-cell setup. This observation indicates that while \method encounters specific challenges with the ENAS dataset, it holds the potential for robust performance across various two-cell datasets. The distinction in \method's effectiveness may be attributed to how the flow encode module's projection strategy is utilized, underscoring the opportunity for further enhancement in the global attention module to better accommodate datasets with two-cell architectures.

\begin{table}[h!]
\vspace{-2mm}
  \centering
  \caption{Kendall's Tau (scaled up by a factor of 100, mean over 9 trials) on the ENAS dataset. In each setting, the best performances are highlighted in \green{green}.
  }
  % \vspace{-1mm}
  \scalebox{0.75}{%
  \renewcommand{\arraystretch}{1.1}
    %\begin{tabular}{l|c|c|c|c|c|c|c|c}
    \begin{tabular}{l|c c c c|c}
        \thickhline
        Datasets& \multicolumn{4}{c|}{ENAS}&Avg.\\
        %\midrule
        %\cmidrule{1-13}
        \cline{1-5}
        Training portions& 1\% & 5\% & 10\% & 50\% & Rank\\
        \hline\hline
        \GCN~\cite{bresson2017residual}
        & 15.0 & 36.1 & 41.2& 54.7 & 4.75\\
        \DAGNN~\cite{thost2021directed}
        & \best 31.0 & \best 47.0& \best 52.6& 61.3 & 1.25\\
        \GPS~\cite{rampavsek2022recipe}
        & 6.9 & 26.5 & 34.2 & 51.2&6.00\\
        \DAGFo~\cite{luo2023transformers}
        & 12.2 & 41.4& 46.5& 57.9 &4.25 \\
       
        \TAGATES~\cite{ning2022ta}
        & 22.9& 45.2& 49.4 & 61.2 & 2.50\\
        \hline\hline
        \method
        & 18.8  & 44.3& 49.5& \best 64.7 & 2.25\\
    \thickhline
    \end{tabular}
    }
    \label{tab:enas}
\end{table}
\color{black}
\label{sec:sup}

\end{document}